\documentclass{article}

\usepackage{arxiv}

\usepackage[utf8]{inputenc} 
\usepackage[T1]{fontenc}    
\usepackage{hyperref}       
\usepackage{url}            
\usepackage{booktabs}       
\usepackage{amsfonts}       
\usepackage{nicefrac}       
\usepackage{microtype}      
\usepackage{lipsum}		
\usepackage{graphicx}
\usepackage{natbib}
\usepackage{amsmath}

\title{Enhancing Text-based Knowledge Graph Completion with Zero-Shot Large Language Models: A Focus on Semantic Enhancement}

\author{ Rui Yang \\
	Sun Yat-sen University\\
	Guangzhou, 510080, China \\
	\texttt{yangr83@mail2.sysu.edu.cn} \\
	\And
        Jiahao Zhu \\
	Sun Yat-sen University\\
	Guangzhou, 510080, China \\
	\And
        Jianping Man \\
	Sun Yat-sen University\\
	Guangzhou, 510080, China \\
	\And
 	Li Fang \\
	Sun Yat-sen University\\
	Guangzhou, 510080, China \\
	\texttt{fangli9@mail.sysu.edu.cn} \\
 	\And
        Yi Zhou \\
	Sun Yat-sen University\\
	Guangzhou, 510080, China \\
	\texttt{zhouyi@mail.sysu.edu.cn} \\
}

\renewcommand{\shorttitle}{}

\begin{document}
\maketitle

\begin{abstract}
The design and development of text-based knowledge graph completion (KGC) methods leveraging textual entity descriptions are at the forefront of research. These methods involve advanced optimization techniques such as soft prompts and contrastive learning to enhance KGC models. The effectiveness of text-based methods largely hinges on the quality and richness of the training data. Large language models (LLMs) can utilize straightforward prompts to alter text data, thereby enabling data augmentation for KGC. Nevertheless, LLMs typically demand substantial computational resources. To address these issues, we introduce a framework termed constrained prompts for KGC (CP-KGC). This CP-KGC framework designs prompts that adapt to different datasets to enhance semantic richness. Additionally, CP-KGC employs a context constraint strategy to effectively identify polysemous entities within KGC datasets. Through extensive experimentation, we have verified the effectiveness of this framework. Even after quantization, the LLM (Qwen-7B-Chat-int4) still enhances the performance of text-based KGC methods \footnote{Code and datasets are available at \href{https://github.com/sjlmg/CP-KGC}{https://github.com/sjlmg/CP-KGC}}. This study extends the performance limits of existing models and promotes further integration of KGC with LLMs.
\end{abstract}


\section{Introduction}
\label{sec:Introduction}
Knowledge graphs (KGs) represent real-world facts as triples and are highly valuable in applications such as question answering \cite{sun2019pullnet} and recommendation systems \cite{huang2018improving}. KGs effectively encapsulate relationships among entities through a triple-based structure. However, despite their wide-ranging applications, KGs often suffer from incompleteness. To address this issue, developing techniques for KG completion (KGC) is crucial for automated construction and validation of KGs.

Current KGC methods are typically categorized into two main types: embedding-based and text-based methods. Embedding-based methods assign a low-dimensional vector to each entity and relationship, functioning without additional information such as entity descriptions. Popular models include TransE \cite{bordes2013translating}, TransH \cite{wang2014knowledge}, and RotatE \cite{sun2019rotate}. Conversely, text-based methods \cite{wang2022simkgc, yao2019kg} integrate entity descriptions into the learning process. These methods feed textual representations of entities and their relationships into a language model, which subsequently calculates a score to assess the validity or plausibility of the triples. As text-based KGC models excel in utilizing detailed, contextual text, they enhance our comprehension of intricate semantic links within KGs. Therefore, this area is currently the subject of intensive research.

\begin{figure*}[h]
  \centering
  \includegraphics[width=\linewidth]{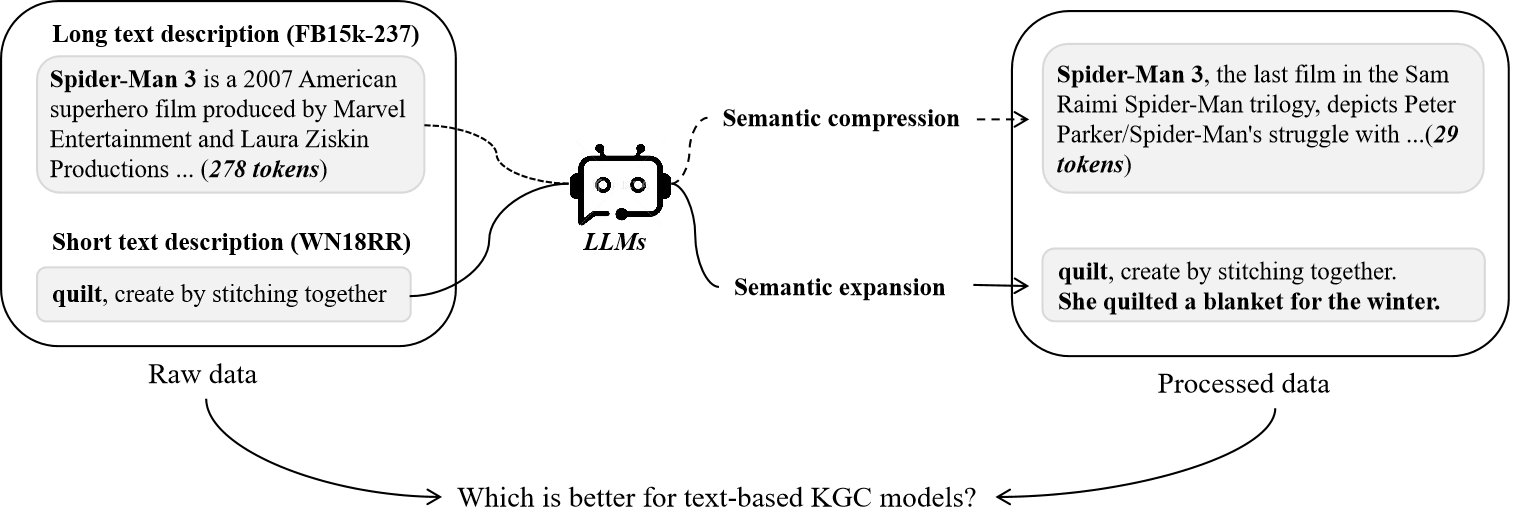}
  \caption{LLMs can add or remove content from entity descriptions.}
  \label{fig:question}
\end{figure*}

However, the performance of text-based KGC models heavily relies on the acquired semantic information. As shown in Figure \ref{fig:question}, we raise the question of whether the texts optimized by LLMs are more effective for text-based KGC models. Pre-trained language models (PLMs) have a predefined maximum input length, which means that when text is inputted, it is truncated when the maximum input length is exceeded. Conversely, shorter texts are padded to reach the maximum length, potentially resulting in less clear and under-detailed semantic representations. Therefore, it is crucial to encapsulate the maximum amount of pertinent information within the truncation threshold. This ensures the maintenance of semantic coherence for enhanced model efficacy.

The advent of large language models (LLMs) has enabled the automatic optimization of KGC data. These models perform significantly advanced various natural language processing tasks using their sophisticated text generation abilities \cite{singhal2023large}. LLMs acquire linguistic patterns and assimilate knowledge from their training data \cite{petroni2019language}. However, prompts are not universally effective and require empirical validation on each specific dataset. The use of LLMs for data augmentation involves some challenges. For example, there is a possibility of inaccuracies in the responses of the model, such as providing irrelevant answers or generating false information. Additionally, employing LLMs for inferential tasks can be computationally demanding. It is crucial to acknowledge that the prompts that effectively optimize text via LLMs and those enhancing text-based KGC models are not universally applicable \cite{amatriain2024prompt}. Effective prompt design for diverse datasets typically requires iterative testing.

To address the above issues, we introduce a framework termed constrained prompts for KGC (CP-KGC), which utilizes LLMs for data enhancement. The CP-KGC framework designs prompts that support semantic compression and expansion. These prompts are designed to enhance semantic richness, irrespective of text length. Additionally, to address content inaccuracies and minimize noise, our prompts incorporate contextual constraints pertinent to the involved entities. We also evaluate LLMs at varying parameter scales, including Qwen-7B-Chat-int4, Qwen-7B-Chat \cite{Jinze2023Qwen}, LLaMA2-7B-Chat \cite{touvron2023llama}, Qwen-turbo, GPT-3.5-turbo, and GPT-4 \cite{achiam2023gpt}. These evaluations offer certain insights for future research on integrating LLMs with KGs. Our principal contributions are as follows:

\begin{itemize}
    \item We validated the effectiveness of CP-KGC using three publicly available datasets. CP-KGC enhanced the semantic integrity of text and further improved the performance of text-based models.
    \item We demonstrated the effectiveness of a quantized model (Qwen-7B-Chat-int4) using LLMs with various parameter scales, enhancing the feasibility of further research on integrating KGC and LLMs.
    \item The contextual constraints strategy of CP-KGC accurately identified polysemous entities in datasets. Additionally, this strategy could enhance the stability of text generation by LLMs.
\end{itemize}

The remainder of this paper is organized as follows: Section \ref{sec:Related Work} introduces prior research pertinent to this study. Section \ref{sec:method} describes the methodology focused on achieving semantic completeness. Section \ref{sec:Experiments} details the experimental results. Section \ref{sec:Analyze} presents the analysis of the stability and contextual functionality contributions of LLMs. Section \ref{sec:conclusion} presents a summary and considerations for future research.

\section{Related Work}\label{sec:Related Work}

\subsection{Knowledge Graph Completion}
KGC techniques aim to improve the comprehensiveness of KGs by tackling the issue of missing connections. In embedding-based methods such as TransE \cite{bordes2013translating}, entities and relations are represented within a low-dimensional vector space, simplifying relationship modeling by translating entity vectors. Building on TransE, methods such as TransH \cite{wang2014knowledge} and TransR \cite{lin2015learning} have been developed. TransH introduces hyperplanes to differentiate entity roles within relations, while TransR separates entity and relation embeddings into distinct spaces, enabling more nuanced interactions. The ComplEx model \cite{trouillon2016complex} enhances embedding expressiveness using complex numbers, effectively capturing symmetric and asymmetric relationships. Similarly, the RotatE \cite{sun2019rotate} model uses rotation angles in a complex vector space to evaluate the similarity between triples, facilitating the representation of relational patterns such as symmetry and inversion. These embedding methods are straightforward and effective for encoding factual information. However, they mainly focus on the triples information within KGs \cite{wang2021kepler}.

Text-based approaches focus on encoding the descriptions of entities. KG-BERT \cite{yao2019kg} is the pioneer in applying PLMs to KGC tasks, representing triples and their textual descriptions as continuous text sequences. The veracity of triples is ascertained using the final hidden state of the $[CLS]$ token, marking a significant innovation in KGC field. To address the high computational demands and limited structured knowledge of the text encoder in KG-BERT, Wang et al.\cite{wang2021structure} proposed StAR that integrates text encoding with graph embedding techniques. This model divides triples into two parts using a Siamese-style text encoder, enhancing structured knowledge and acquiring contextualized representations. Subsequently, Chen et al.\cite{chen2022knowledge} introduced the KG-S2S framework: a sequence-to-sequence PLM approach that uniformly addresses various structured tasks in KGC. This framework transforms KG facts into flat texts and utilizes entity descriptions and soft prompts, improving performance across different benchmark datasets. Additionally, Chen et al.\cite{chen2023dipping} introduced CSProm-KG, which merges structural and textual information of KGs using conditional soft prompts, thereby enhancing the effectiveness of KGC tasks.

These methods are generally less effective than embedding-based approaches. Wang et al.\cite{wang2022simkgc} introduced SimKGC, which improves text-based methods for KGC tasks by integrating PLMs with contrastive learning techniques. However, text-based methods focus on model optimization, and their performance is further influenced by the semantic completeness that PLMs can capture.

\subsection{Large Language Models for knowledge graphs}
Recent LLMs such as Qwen \cite{Jinze2023Qwen}, Llama2 \cite{touvron2023llama}, and GPT-4 \cite{achiam2023gpt}, known for their emergent capabilities \cite{zhao2023survey,wei2022emergent}, have significantly advanced the use of LLMs in improving KGs \cite{pan2024unifying}. These models can generate new facts for KG construction using simple prompts \cite{zhou2022large,shin2020autoprompt,brown2020language,jiang2020can}. Moreover, the introduction of chain-of-thought techniques has significantly advanced the reasoning abilities of LLMs. Wei et al.\cite{wei2022chain} introduced chain-of-thought prompting to improve model performance on arithmetic, common sense, and symbolic reasoning tasks by incorporating intermediate reasoning steps in prompts.

Currently, most LLMs are built upon the transformer architecture \cite{pan2024unifying, vaswani2017attention}. This architecture also underpins text-based KGC methods. For example, SimKGC \cite{wang2022simkgc} is based on BERT \cite{devlin2018bert}, StAR \cite{wang2021structure} is built on RoBERTa \cite{liu2019roberta}, and KG-S2S \cite{chen2022knowledge} is developed using T5 \cite{raffel2020exploring}. The emergence of LLMs has provided a new direction for KGC research. Traditional summarization methods compress semantic content but often fail to expand the semantics of overly concise texts \cite{bano2023summarization, bano2022bert}. By contrast, the use of LLMs for semantic expansion tasks enables the acquisition of more comprehensive semantic information through prompts.

Herein, we propose a text semantic enhancement framework called CP-KGC. CP-KGC utilizes entities and their descriptions as contextual information to constrain LLM outputs by adhering to standard conversational patterns. Furthermore, CP-KGC can enhance the semantic integrity of text regardless of the length of entity descriptions.

\section{Methodology}\label{sec:method}
\begin{figure*}[h]
  \centering
  \includegraphics[width=\linewidth]{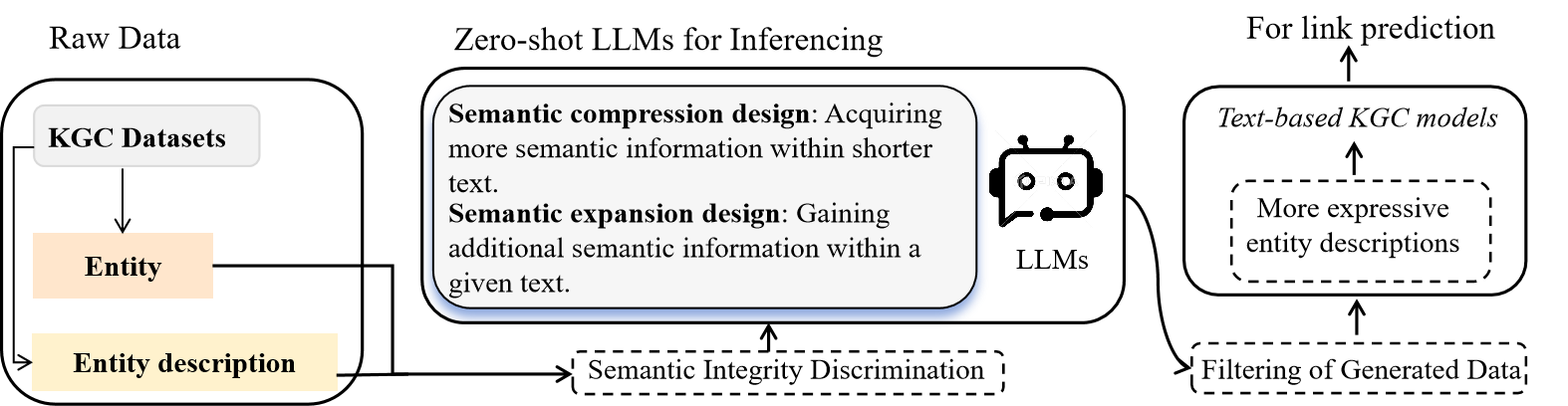}
  \caption{CP-KGC semantic enhancement framework.}
  \label{fig:cpkgc}
\end{figure*}

In this section, we introduce the CP-KGC framework, which comprises several components: semantic integrity discrimination, zero-shot reasoning with LLMs, filtering of generated data, and re-prediction using text-based methods. The main framework is illustrated in Figure \ref{fig:cpkgc}.

\subsection{Semantic Integrity Discrimination}
In this paper, we focus on enhancing semantic information within constrained text lengths to bolster text-based methods leveraging PLMs such as BERT or T5. Initially, we define each entity $e \in E$, comprising the entity name $e$ and its description $desc_e$. $E$ denotes the set of entities in KGs. The total text length of an entity, $L_e$, is determined by the following equation:
\begin{equation}
L_e = len(e) + len(desc_e),
\end{equation}
where $len$ denotes the function for calculating text length, employing the same tokenization method as used in BERT.

It is necessary to determine whether the text length of an entity exceeds the maximum truncation length of the model: $L_\text{max}$. For $L_e > L_\text{max}$, we used a strategy of semantic compression to fit the length within the input constraints of the model without sacrificing essential information, as detailed in Section \ref{sec:section3.2}. Conversely, for $L_e < L_\text{max}$, we employed a strategy of semantic expansion to augment the entity description, thereby introducing additional contextual information, as detailed in Section \ref{sec:section3.3}.

\subsection{Semantic Compression}\label{sec:section3.2}
We designed CP-KGC to enhance the semantic integrity of $desc_e$ using simple prompts. In our semantic compression experiments, we employed the FB15k-237 dataset, featuring entity descriptions in an extensive narrative style with an average length of 139 tokens. Although fully detailed texts offer more semantic content, they significantly increase resource demands.

We employed zero-shot LLMs to explore the impact of complete semantic information on the model performance. LLMs might be affected by polysemy when generating textual descriptions directly from entity; therefore, we designed the following prompt:

\begin{description}
  \item[$\textit{desc}_e$ =] \textit{'description for entity';}
  \item[$e$ =] \textit{'entity name';}
\end{description}

\noindent
\textbf{Prompt for semantic compression:}
\begin{itemize}
  \item \textit{\textbf{$\textit{desc}_e$} is the description of the $\textbf{e}$. Please summarize \textbf{$\textit{desc}_e$} in one sentence as briefly as possible:}
\end{itemize}

This prompt uses $\textit{desc}_e$ as a constraint to prevent LLMs from generating irrelevant content, which helps minimize polysemy-related issues. The summary within the prompt focuses on the entity, ensuring greater accuracy in the generated content and preventing the omission of crucial information. A cleaning process is required for texts produced through the semantic compression prompt: 
\begin{equation}
cleaned_e = h(prompt_\textit{compressed}(e, desc_e))
\end{equation}
$h$ denotes the cleaning function, where $\textit{prompt}_\textit{compressed}$ represents using the above prompt to guide LLMs in generating new text. The content generated by LLMs may include conversational elements or fail to provide answers. We employed regular expressions and keyword matching to cleanse these portions of the text. $cleaned_e$ denotes the target text after its final cleaning and filtration.

The majority of existing text-based approaches are trained based on PLMs. Let the length of $cleaned_e$ be denoted as $T$, and let $L_\textit{max}$ represent the maximum truncation length for PLMs. If $ T > L_{\textit{max}} $, then $cleaned_e$ must be trimmed to produce the sequence $cleaned_e'$, where:
\begin{equation}
cleaned_e' = cleaned_e[ : L_{\textit{max}} ]
\end{equation}
$\textit{cleaned}_e'$ comprises the initial $ L_{\textit{max}} $ tokens of $\textit{cleaned}_e$. Following this, we employed the standard input processing techniques associated with PLMs.

\begin{equation}
\textit{input}_{e_h, e_t} = \textit{[CLS]}\textit{cleaned}_{e_h}'\textit{[SEP]}\textit{relation}\textit{[SEP]}\textit{cleaned}_{e_t}'\textit{[SEP]}
\end{equation}
\begin{equation} 
f(\textit{input}_{e_{h_i}, e_{t_i}}), \quad \forall (e_{h_i}, e_{t_i}) \in E \times E 
\end{equation}

Here, $\textit{[CLS]}$ is a special token used at the beginning of the input to indicate the start of a sequence and $\textit{[SEP]}$ is a separator token used to demarcate different segments within the input. $e_h$ and $e_t$ denote the head and tail entities in KGs, respectively, and $i$ refers to instances of $e_h$ and $e_t$ in the dataset. $f$ represents a text-based KGC model, and the data processed as described above can be directly used in various models. This format helps the PLM understand the input components as distinct but related elements.

\subsection{Semantic Expansion}\label{sec:section3.3}
The use of semantic compression allows PLMs to receive relatively complete semantic information. Although LLMs still lose some semantic information when summarizing text, this approach effectively optimizes longer texts. Sometimes, entity descriptions are too short to fully represent their meanings. To investigate whether further extension of descriptions would be effective, we conducted experiments on the WN18RR and FB15k-237 datasets.

The WN18RR dataset does not describe entities with long texts. For example, the description of “quilt” is “stitch or sew together; quilt the skirt” and for “raise,” it is “move upwards; lift one’s eyes.” These descriptions are used to define the entities. Entity description focuses on various relationships among word meanings, parts of speech, and words. In the original entity descriptions of WN18RR, some words have usage examples. For instance, the description of “tinsel-VB-2” is “adorn with tinsel; snow flakes tinseled the trees.” Therefore, we considered generating usage examples for entities with shorter textual descriptions to expand their semantic information. In FB15k-237, we supplemented the existing text descriptions based on the original data.

We designed the following prompts for semantic expansion:

\begin{description}
  \item[$\textit{desc}_e$ =] \textit{'description for entity';}
  \item[$e $ =] \textit{'entity name';}
\end{description}
\noindent
\textbf{Prompt for semantic expansion(WN18RR): }
\begin{itemize}
  \item \textit{\textbf{$e$} means \textbf{$\text{desc}_e$}, please use the shortest possible text to introduce the usage of \textbf{$e$}.}
\end{itemize}
\noindent
\textbf{Prompt for semantic expansion(FB15k-237): }
\begin{itemize}
  \item \textit{Please regenerate the description of \textbf{$e$} based on \textbf{$\text{desc}_e$}. You just need answer the regenerated text description!}
\end{itemize}

We retained the original descriptions of entities in the semantic expansion prompt. This strategy reduces redundant text because of polysemy and decreases the likelihood of erroneous or hallucinatory outputs. Unlike the compression prompt, the semantic expansion prompt is designed to elicit texts with enriched semantic content. Implementing constraints is essential to deter the model from regurgitating existing content. Following the regeneration of the text, it proceeds as follows:

\begin{equation}
cleaned_e = h(e + \textit{desc}_e + \textit{prompt}_\textit{expansion}(e + \textit{desc}_e))
\end{equation}

Similar to Section \ref{sec:section3.2}, $h$ represents the cleaning function. However, by contrast, this function integrates the entity with its original description to produce a cleaned and expanded text. Subsequently, equations (3)–(5) are reapplied to retrain the model.

\section{Experiments}\label{sec:Experiments}

\subsection{Dataset}

We conducted evaluations using three common datasets: FB15k-237, WN18RR, and UMLS. Table \ref{tab:dataset} shows their statistical results. FB15k-237, a benchmark dataset for KGC, is derived from the original Freebase knowledge base. It excludes inverse relations to prevent test leakage. WN18RR, based on WordNet and utilized for evaluating relationship and link prediction models, removes similar inverse relations from the original WN18, comprising ~110k triples and over 40k entities. UMLS is a medical semantic network encompassing semantic types (entities) and relationships, focusing on medical data integrity and applicability.

\begin{table}[htbp]
  \centering
  \caption{The statistics of the datasets utilized in this paper.}
  \label{tab:dataset}
    \begin{tabular}{cccccc}
    \hline
    dataset & \#entity & \#relation & \#train & \#valid & \#test \\
    \hline
    UMLS & 135 & 46 & 5216 & 652 & 661 \\
    WN18RR & 40943 & 11 & 86835 & 3034 & 3134 \\
    FB15k-237 & 14541 & 237 & 272115 & 17535 & 20466 \\
    \hline
    \end{tabular}
\end{table}

\subsection{Experimental Setup}
\subsubsection{Model Selection}
We chose SimKGC as the base model for CP-KGC. This model outperforms other text-based approaches in efficiency and overall performance. To comprehensively evaluate the variation in the performance of LLMs based on model parameter sizes, we utilized LLMs including GPT-4, GPT-3.5-turbo, Qwen-turbo, LLaMA2-7B-Chat, Qwen-7B-Chat, and Qwen-7B-Chat-int4. We conducted all model inference and training tasks on a single A800-80GB GPU, ensuring a consistent computational environment. Moreover, we employed multiple GPUs to enhance the efficiency of the validation process. We utilized a subscription-based API service for the larger models, such as GPT-4, GPT-3.5-turbo, and Qwen-turbo, providing reliable access and scalability for our experiments. This setup allowed us to accurately assess the scalability and performance implications of various LLM configurations under controlled conditions.
 
\subsubsection{Evaluation Metrics}
Building upon prior research, we implemented the CP-KGC method and incorporated text-based strategies for extended predictions with augmented data. We assessed our model using four standardized evaluation metrics: mean reciprocal rank (MRR) and Hits@k (k = 1, 3, and 10). MRR, specifically tailored for KGC tasks, gauges the performance of the model by computing the reciprocal rank of the first accurate answer for each query and averaging these reciprocals across all queries. Hits@k metrics evaluate whether the correct answer falls within the top k predictions made by the model. If the correct answer ranks among the top k predictions, Hits@k is 1; otherwise, it is 0. MRR and Hits@k are calculated within a filtering framework, where scores associated with all known true triples from the training, validation, and test sets are disregarded. These metrics are averaged across two dimensions: head entity and tail entity predictions.

\subsection{Main Result}
We have summarized the main results in Table \ref{tab:main results}, evaluating those using commonly used link prediction indicators. In the remaining part of this section, we elaborate on the results of semantic compression and semantic expansion prompts.
\begin{table}[htbp]
  \centering
  \caption{Main results for WN18RR and FB15k-237 datasets. The notion w refer to the use of the CP-KGC framework. The LLM employed by CP-KGC in this table is GPT-3.5-turbo. The bolded numbers indicate that the results improved after optimizing the CP-KGC.}
  \label{tab:main results}
  \resizebox{\textwidth}{!}{%
  \begin{tabular}{l|cccc|cccc}
    \toprule
     & \multicolumn{4}{c|}{WN18RR} & \multicolumn{4}{c}{FB15k-237} \\
    \cmidrule(lr){2-5} \cmidrule(lr){6-9}
    Method & MRR & Hits@1 & Hits@3 & Hits@10 & MRR & Hits@1 & Hits@3 & Hits@10 \\
    \midrule
    \multicolumn{9}{l}{\textit{embedding-based methods}}\\
    \hline
    TransE\cite{bordes2013translating} & 24.3 & 4.3 & 44.1 & 53.2 & 27.9 & 19.8 & 37.6 & 44.1 \\
    DistMult\cite{yang2014embedding} & 44.4 & 41.2 & 47.0 & 50.4 & 28.1 & 19.9 & 30.1 & 44.6 \\
    ConvE\cite{dettmers2018convolutional} & 45.6 & 41.9 & 47.0 & 53.1 & 31.2 & 22.5 & 34.1 & 49.7 \\
    RotatE\cite{sun2019rotate} & 47.6 & 42.8 & 49.2 & 57.1 & 33.8 & 24.1 & 37.5 & 53.3 \\
    \hline
    \multicolumn{9}{l}{\textit{text-based methods}}\\
    \hline
    KG-BERT\cite{yao2019kg} & 21.6 & 4.1 & 30.2 & 52.4 & - & - & - & 42.0 \\
    StAR\cite{wang2021structure} & 40.1 & 24.3 & 49.1 & 70.9 & 29.6 & 20.5 & 32.2 & 48.2 \\
    CSProm-KG\cite{chen2023dipping} & 57.5 & 52.2& 59.6& 67.8& 35.8& 26.9& 39.3& 53.8\\
    \hline
    KG-S2S\cite{chen2022knowledge} & 57.4 & 53.1 & 59.5 & 66.1 & 32.6 & 24.9 & 35.8 & 48.7 \\
    w CP-KGC & \textbf{57.9} & \textbf{53.3} & \textbf{60.3} & \textbf{66.7} & \textbf{32.8} & \textbf{25.1} & \textbf{36.2} & \textbf{48.9} \\
    SimKGC\cite{wang2022simkgc} & 63.8 & 56.3 & 67.8 & 77.6 & 32.5 & 23.9 & 35.0 & 49.6 \\
    w CP-KGC & \textbf{64.8} & \textbf{58.0} & \textbf{68.3} & 77.3 & \textbf{32.9} & \textbf{24.3} & \textbf{35.3} & \textbf{50.3} \\
    \bottomrule
  \end{tabular}}
\end{table}
\subsubsection{Results of Semantic Compression} 
We conducted the semantic compression experiment on the FB15k-237 dataset with a maximum truncation length of 30 tokens. Under the same conditions as SimKGC, the MRR, Hits@1, Hits@3, and Hits@10 improved by 0.4\%, 0.4\%, 0.3\%, and 0.7\%, respectively, with an average improvement across all metrics of 0.45\%. We also verified the results of CP-KGC on KG-S2S. Using simple prompt templates for generating concise texts, we successfully surpassed the performance thresholds of the original models without modifying additional parameters.

Although a modest enhancement in performance metrics occurs after semantic compression, the extent of this improvement is minimal, with the greatest increase being a 0.7\% improvement in SimKGC (Hits@10 in FB15k-237). This limited enhancement is attributed to several factors. First, the occurrence of entities lacking textual descriptions during training created gaps that impacted prediction outcomes. Second, the deliberate removal of many inverse relationships in FB15k-237 complicated the task. Relationships that could previously be inferred through simple logical deductions now required more complex paths and reasoning using the models. Finally, despite using contextual constraints to focus prompts on entities, the semantic compression process may still lead to semantic information loss.

\begin{figure*}[h]
  \centering
  \includegraphics[width=\linewidth]{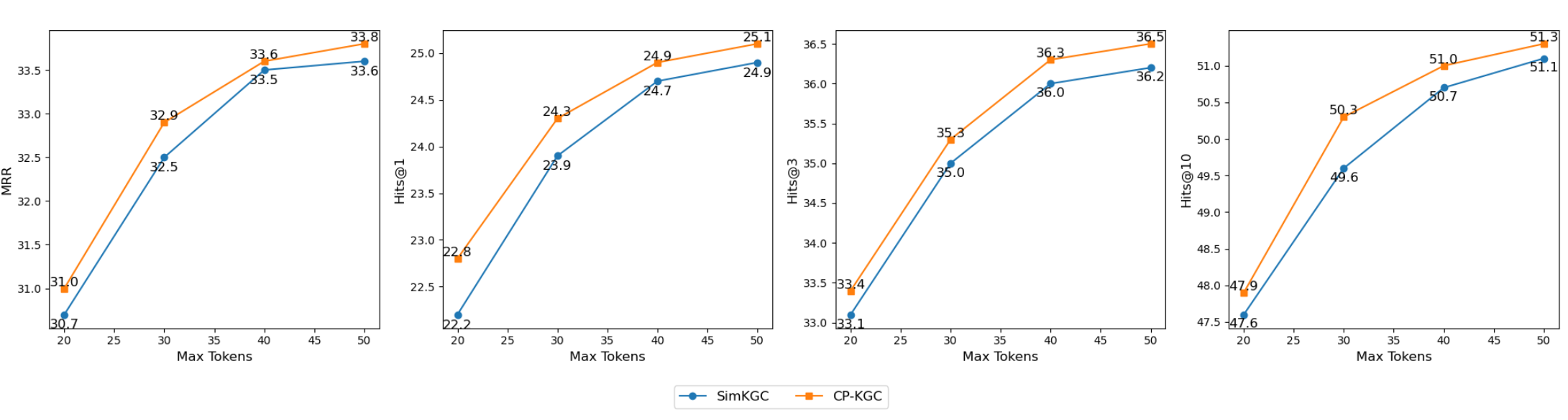}
  \caption{Comparing the impact of different maximum truncation lengths on model performance between CP-KGC and SimKGC.}
  \label{fig:length}
\end{figure*}

To investigate the impact of text truncation length on semantic retrieval, we analyzed the role of semantic compression by adjusting the maximum truncation length. Figure \ref{fig:length} shows that the maximum text truncation length is set between 20 and 50 to compare the results between texts enhanced by CP-KGC and the original texts. The performance of the model exhibited a positive correlation with maximum truncation, although this trend diminished as the truncation length increased. This suggests that the performance of the model can be enhanced to a certain extent based on the richness of the semantic information expressed by entities. Additionally, under identical truncation conditions, texts optimized using CP-KGC demonstrated superior results. This indicates that acquiring more semantic information within the shortest possible text length enhances model performance. This finding supports the efficiency of CP-KGC in capturing essential semantic content.

\subsubsection{Results of Semantic Expansion}\label{sec:Results on semantic expansion}
Table \ref{tab:main results} shows the results of semantic expansion using the WN18RR dataset. These results show that CP-KGC led to more substantial improvements for SimKGC compared with the FB15k-237 dataset. SimKGC achieved an enhancement with a 1\% increase in MRR and a 1.7\% increase in Hits@1, while Hits@10 exhibited a slight decline. This decline in Hits@10, a more lenient metric than Hits@1 and Hits@3, is probably due to its broader inclusion criteria, allowing a correct answer within the top 10 predictions. As such, the enhanced textual richness does not significantly boost the precision for Hits@10. Regardless of the detail level of the original text, the model effectively identifies the correct answer within the top predictions.

Unlike the FB15k-237 dataset, which mainly includes names of individuals and locations, the WN18RR dataset organizes entities into four grammatical categories, i.e., nouns, verbs, adjectives, and adverbs, with a predominant focus on nouns and verbs. Typically, succinct descriptions and illustrative examples are sufficient to clarify entities meaning. Excessively elaborate expressions can introduce noise and diminish model performance, particularly in datasets featuring complex grammatical categorizations.

\begin{table}[htbp]
  \centering
  \caption{Results of SimKGC and CP-KGC on FB15k-237 and FB15k-237N with a model truncation length set to 130. The notion w refer to the use of the CP-KGC framework. The bolded numbers indicate that the results improved after optimizing the CP-KGC.}
  \label{tab:15and15k}
  \resizebox{\textwidth}{!}{%
  \begin{tabular}{l|cccc|cccc}
    \toprule
     & \multicolumn{4}{c|}{FB15k-237} & \multicolumn{4}{c}{FB15k-237N} \\
    \cmidrule(lr){2-5} \cmidrule(lr){6-9}
    Method & MRR & Hits@1 & Hits@3 & Hits@10 & MRR & Hits@1 & Hits@3 & Hits@10 \\
    \midrule
    SimKGC & 34.5 & 25.6 & 37.4 & 52.2 & 37.2 & 28.9 & 40.2 & 53.4 \\
    w CP-KGC & \textbf{34.7} & \textbf{26.0} & \textbf{37.5} & \textbf{52.3} & \textbf{37.5} & \textbf{29.2} & \textbf{40.3} & 53.4\\
    \bottomrule
  \end{tabular}}
\end{table}

The FB15k-237 and FB15k-237N datasets share information on the entity description. Table \ref{tab:15and15k} shows the results when the truncation length is 130. The results show that CP-KGC can further improve the performance of the model. However, the improvement are significant when these results are contrasted with those from shorter truncation lengths presented in Table \ref{tab:main results}. This suggests that accurate entity representation requires text rich in semantic information. The amount of required semantic information varies based on entity type; nouns and verbs need only simple examples, while names of individuals and places required more detailed semantic content. It is important to use a large batch size of 1024 to effectively demonstrate the benefits of contrastive learning in SimKGC. Additionally, a truncation length of 130 necessitates 246 GB of GPU memory.

\subsubsection{Experimental Results on UMLS}\label{sec:Results on UMLS}
To further validate the effectiveness of CP-KGC, we selected the widely used KGC dataset UMLS for additional testing. Notably, previous studies on this dataset primarily focused on reporting only the Hits@10 and Mean Rank metrics. To deepen our understanding and provide a comprehensive performance assessment, we retested the models KG-BERT and SimKGC.

In the UMLS dataset, entity descriptions originally comprised an average of 212 tokens in length. We used a semantic compression prompt to maintain semantic integrity while optimizing data processing. This approach effectively reduces the average description length to 19 tokens. These shortened texts are then reintroduced into our models for further testing. The results of this experimentation, assessing the impact of description condensation on model performance, are detailed in Table \ref{tab:umls}.

\begin{table}[htbp]
  \centering
  \caption{CP-KGC demonstrated marked improvements in performance on the KG-BERT and SimKGC models. The notion w refer to the use of the CP-KGC framework. The LLM employed by CP-KGC in this table is GPT-3.5-turbo. The bolded numbers indicate that the results improved after optimizing the CP-KGC.}
  \label{tab:umls}
  \begin{tabular}{lcccc}
    \toprule
     & \multicolumn{4}{c}{UMLS} \\
    \cmidrule(lr){2-5}
    Method & MRR & Hits@1 & Hits@3 & Hits@10 \\
    \midrule
    KG-BERT & 64.8 & 53.9 & 71.4 & 83.9  \\
    w CP-KGC & \textbf{79.8} & \textbf{67.6} & \textbf{89.7} & \textbf{98.0} \\
    SimKGC & 68.8 & 57.9 & 74.8 & 89.4  \\
    w CP-KGC & \textbf{78.0} & \textbf{67.8} & \textbf{85.7} & \textbf{95.1} \\
    \bottomrule
  \end{tabular}
\end{table}

Semantic compression has allowed KG-BERT and SimKGC to produce texts with enhanced semantic completeness, facilitating a more accurate interpretation of entity meanings by these models. In KG-BERT, CP-KGC achieved an 18.3\% increase in Hits@3, with an overall average improvement of 15.3\% across all metrics. CP-KGC also showed effectiveness in SimKGC, with an average performance improvement of 8.9\%.

In the UMLS dataset, the performance gains of CP-KGC were notably more pronounced than those on FB15k-237 and WN18RR datasets. We believe the reasons for this observation are as follows: on the one hand, the original descriptions in UMLS were lengthy and semantically diverse, mostly consisting of conceptual entities; on the other hand, the smaller size of the UMLS dataset tended to highlight differences in performance metrics more clearly.

\section{Analysis}\label{sec:Analyze}
\subsection{Comparison of LLMs}
The capabilities of LLMs generally correlate with their parameter sizes. Models with larger parameters require more computational resources \cite {kaplan2020scaling}. The tasks of semantic compression and expansion are relatively simple. We conducted experiments using models of different parameter sizes to study the combination of KGC tasks and LLMs under hardware resource constraints. We generated new text using different LLMs. Table \ref{tab:compare} shows the detailed results.

In the WN18RR dataset, LLMs showed enhanced performance for MRR, Hits@1, and Hits@3 but a minor decline in the performance for Hits@10. We discussed the potential reasons for this trend in Section \ref{sec:Results on semantic expansion}.

\begin{table}
\caption{Main results for WN18RR and FB15k-237 datasets with different LLMs. The notion w refer to the use of the CP-KGC framework. The bolded numbers indicate that the results improved after optimizing the CP-KGC.}
\label{tab:compare}
\resizebox{\textwidth}{!}{%
\begin{tabular}{c|cccc|ccccl}
\toprule
& & WN18RR & & & & &FB15K237 & \\
Method & MRR & Hits@1 & Hits@3 & Hits@10 & MRR & Hits@1 & Hits@3 & Hits@10 \\
\midrule
\texttt{SimKGC} & 63.8 & 56.3 & 67.8 & 77.6 & 32.5 & 23.9 & 35.0 & 49.6 \\
\texttt{w Qwen-7B-Chat} & \textbf{64.6} & \textbf{57.5} & \textbf{68.6} & \textbf{77.7} & \textbf{33.0} & \textbf{24.6} & \textbf{35.5} & \textbf{50.1} \\
\texttt{w LLaMA2-7B-Chat} & \textbf{64.6} & \textbf{57.4} & \textbf{68.4} & 77.5 & \textbf{32.6} & \textbf{24.1} & \textbf{35.1} & \textbf{49.8} \\
\texttt{w Qwen-7B-Chat-int4} & \textbf{64.7} & \textbf{57.7} & \textbf{68.4} & 77.4 & \textbf{32.8} & \textbf{24.2} & \textbf{35.2} & \textbf{50.1} \\
\texttt{w Qwen-turbo} & \textbf{64.8} & \textbf{57.7} & \textbf{69.0} & 77.5 & \textbf{32.9} & \textbf{24.3} & \textbf{35.3} & \textbf{50.3} \\
\texttt{w GPT3.5-turbo} & \textbf{64.8} & \textbf{58.0} & \textbf{68.3} & 77.3 & 32.5 & \textbf{24.0} & 35.0 & \textbf{49.8} \\
\texttt{w GPT4} & \textbf{64.2} & \textbf{57.1} & \textbf{68.0} & 77.1 & \textbf{32.6} & \textbf{24.1} & \textbf{35.2} & \textbf{49.8} \\
\bottomrule
\end{tabular}
}
\end{table}

This study primarily employed semantic compression and expansion to illustrate the general capabilities of LLMs. For example, semantic compression resembles summarization, while semantic expansion involves the extraction and utilization of knowledge. Moreover, during text regeneration with LLMs, we imposed contextual constraints to restrict the outputs, reducing variability among different LLMs. LLMs in the FB15k-237 dataset exhibited enhancements across various metrics, although less prominent than those observed in the WN18RR dataset. The Qwen-7B-Chat-int4 model positively affected semantic compression and expansion tasks, requiring only 10 GB of GPU memory.

To further investigate the stability of text generation by LLMs, we performed additional experiments, which are detailed in Section \ref{sec:Stability}.

\subsection{Stability of LLMs}\label{sec:Stability}
We demonstrated the stability of LLM outputs from two perspectives: the influence of the temperature parameter on the results produced by LLMs and the effective response rate.

\subsubsection{Effect of the Temperature Parameter}

When using API calls for LLMs, the resulting content is significantly affected by the temperature parameter, which is crucial for regulating the diversity of the output text. This parameter typically varies between 0 and 1, occasionally extending to 2 as observed with the Qwen-turbo model, indicating the degree of diversity in the generated text.

As the temperature parameter approaches 0, the resulting text becomes more conservative and predictable. Conversely, as the parameter approaches 1 (or 2 for the Qwen-turbo model), the generated text becomes notably more diverse and inventive. However, it is important to acknowledge that this heightened diversity may also undermine the coherence and accuracy of the output.

\begin{table}[htbp]
  \centering
  \caption{The impact of Qwen-turbo generated text on SimKGC under different temperature parameters. The notion w refer to the use of the CP-KGC framework. The bolded numbers indicate that the results improved after optimizing the CP-KGC.}
  \label{tab:temperature}
  \begin{tabular}{lcccc}
    \toprule
     & \multicolumn{4}{c}{FB15K237} \\
    \cmidrule(lr){2-5}
    Method & MRR & H@1 & H@3 & H@10 \\
    \midrule
    SimKGC & 32.5 & 24.0 & 35.0 & 49.6  \\
    w Qwen-turbo(t=0.5) & \textbf{32.9} & \textbf{24.3} & \textbf{35.3} & \textbf{50.3} \\
    w Qwen-turbo(t=1.0) & \textbf{32.8} &  \textbf{24.2} &  \textbf{35.5}&  \textbf{50.2}\\
    w Qwen-turbo(t=1.5) & \textbf{32.8} & \textbf{24.1} & \textbf{35.4} & \textbf{50.2} \\
    \bottomrule
  \end{tabular}
\end{table}

Taking Qwen-turbo as an example, we tested three temperature parameters (ranging from 0 to 2) on the FB15k-237 dataset, as shown in Table \ref{tab:temperature}.

SimKGC demonstrates consistent performance in evaluating the text generated by Qwen-turbo across various temperature parameters. This is mainly because text-based KGC methods such as SimKGC heavily depend on PLMs, which enhance their comprehension of language overall. These models can capture the underlying meaning of text rather than its superficial literal expressions. For example, BERT is engineered to understand intricate semantic details in text, enabling it to apprehend the intent and content of sentences or paragraphs beyond mere word combinations.

The presented data exhibit minimal fluctuations, with the variance of each metric approaching zero. This consistency across datasets suggests that varying the temperature settings of the same LLM has a negligible effect on the performance of KGC tasks. Furthermore, we posited that the contextual constraints in the prompt limit the variability of content produced by LLMs. This ensures that alterations in the temperature parameter do not substantially deviate the content from the provided context. Consequently, although the text generated under different temperature parameters may appear superficially different, it maintains core meanings or analogous concepts. 

\subsubsection{Effective Answer Generation by LLMs}
We applied filtering functions when utilizing texts generated by LLMs. While we employed contextual constraints to restrict the model output, proving effective in semantic compression on FB15k-237, challenges arise in semantic expansion on WN18RR. For instance, LLMs may encounter difficulties in understanding questions or fail to deliver precise responses. Table \ref{tab:stability} documents the performance of various LLMs.

\begin{table}[h]
  \centering 
  \caption{Generation stability of different LLMs on WN18RR dataset.}
  \label{tab:stability}
  \begin{tabular}{cc}
    \toprule
    Model & Proportion of effective responses \\
    \midrule
    \texttt{Qwen-7B-Chat-int4} & 99.0\% \\
    \texttt{Qwen-7B-Chat} & 99.1\% \\
    \texttt{LLaMA2-7B-Chat} & 98.7\% \\
    \texttt{Qwen-turbo} & 99.7\% \\
    \texttt{GPT3.5-turbo} & 99.7\% \\
    \texttt{GPT4} & 99.8\% \\
    \bottomrule
  \end{tabular}
\end{table}

During response generation, models with larger parameter sizes can tackle a broader spectrum of questions. However, models such as Qwen-7B-Chat and LLaMA2-7B-Chat might generate extraneous content, necessitating the use of supplementary filtering scripts to remove it.

\subsection{The Impact of Context Constraints}

We believe that incorporating contextual constraint strategies into CP-KGC yields more benefits than drawbacks. This assertion has been supported by analyzing two case studies utilizing the WN18RR and FB15k-237 datasets.

\begin{figure*}[h]
  \centering
  \includegraphics[width=\linewidth]{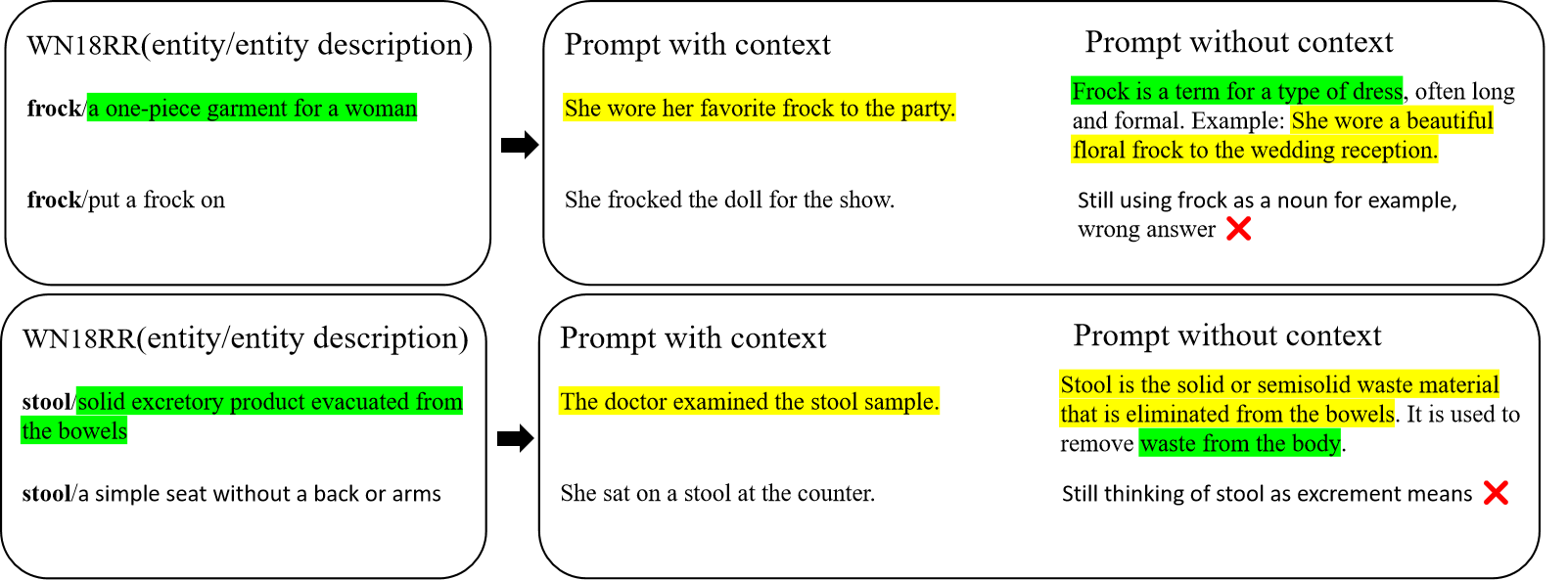}
  \caption{The impact of using context on the WN18RR dataset is illustrated as follows: the yellow section indicates that the predicted part-of-speech usage closely resembles typical usage patterns, whereas the green section shows redundancy with the original entity descriptions.}
  \label{fig:example wn18rr}
\end{figure*}

The WN18RR dataset demonstrates the diverse usages of words, with individual words having multiple meanings. Notably, entities with multiple interpretations account for 31.4\% of the dataset. For example, Figure \ref{fig:example wn18rr} shows that “frock” can serve as a verb and a noun, each conveying distinct meanings. Normally, “frock” is used as a noun, as indicated by the examples with yellow backgrounds provided by the LLM, accurately reflecting its usage. The green background represents entity interpretations provided by the LLM, which are semantically aligned with the original context of the entity. However, when “frock” functions as a verb, the LLM incorrectly treats it as a noun, revealing a challenge in its capacity to manage context-sensitive semantic variations.

\begin{figure*}[h]
  \centering
  \includegraphics[width=\linewidth]{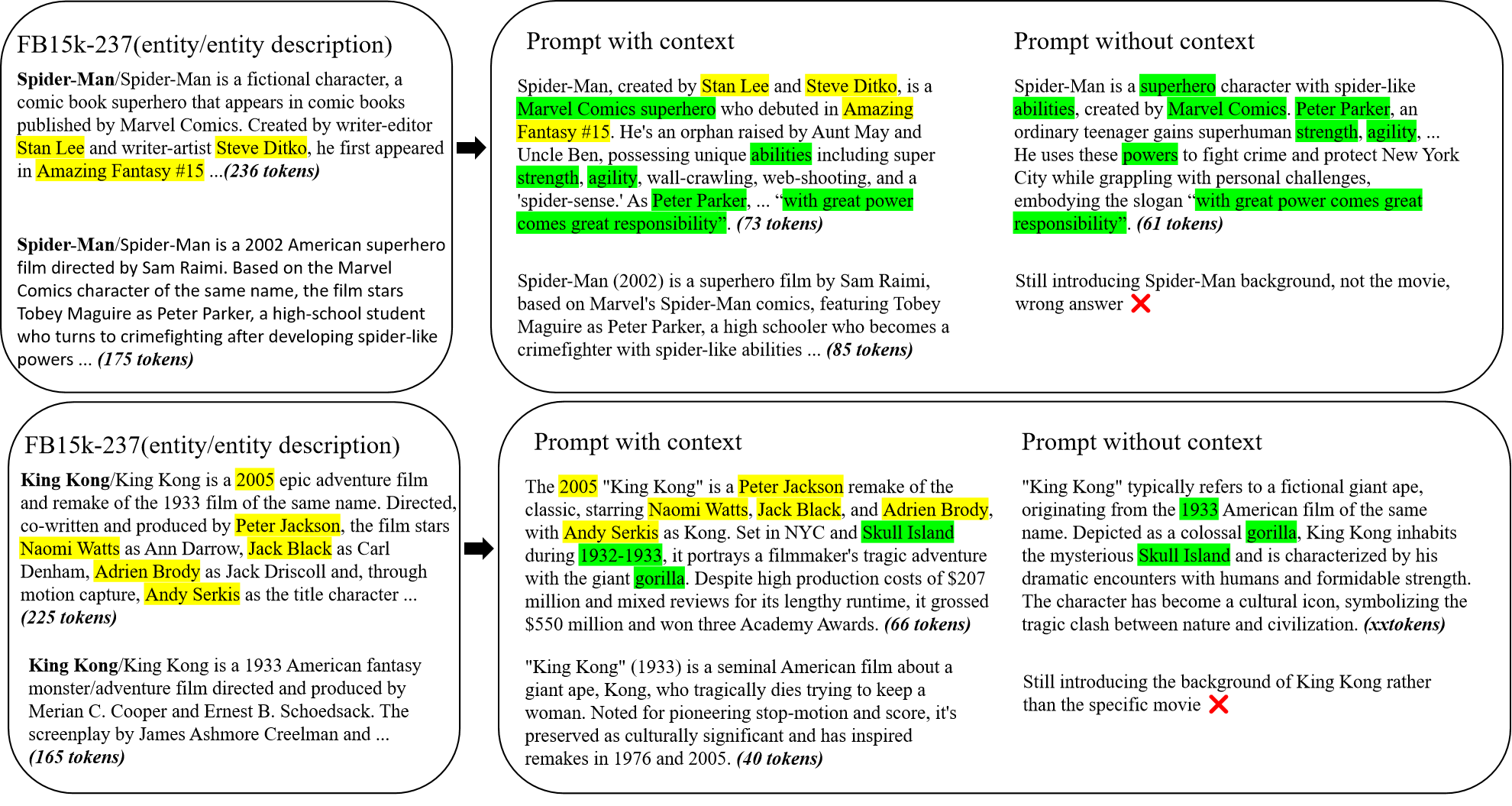}
  \caption{Regarding the FB15k-237 dataset, the influence of context is demonstrated by the yellow section, which highlights fine-grained words introduced by the context, and the green section, which indicates areas where predictions overlap with existing data.}
  \label{fig:example fb15k237}
\end{figure*}

Likewise, the FB15k-237 dataset contains entities with polysemy, constituting 3.5\% of the total dataset. Figure \ref{fig:example fb15k237} shows that the term “Spider-Man” encompasses two distinct entities: one predominantly relates to the character background and the other signifies a film. LLMs can produce pertinent information for both interpretations, irrespective of contextual usage, as demonstrated by the overlapping text highlighted in green.

The use of contextual cues enhances the specificity of the generated content, as demonstrated by the yellow background text, providing details on related characters and comic book issues. Nevertheless, when “Spider-Man” is interpreted in its cinematic context, the LLM tends to default to the most likely interpretation. This often leads to a significant divergence from the intended answer, making the response incorrect. This underscores a crucial challenge in adapting LLMs to accurately capture nuanced distinctions within dataset entities.

Introducing additional context into prompts undoubtedly reduces the diversity of LLMs outputs. However, this approach enables LLMs to produce precise responses in case of words with multiple meanings, thereby minimizing the inclusion of irrelevant or extraneous data.

\section{Conclusion}\label{sec:conclusion}

In this study, we introduce a data augmentation framework named CP-KGC that focuses on semantic completeness. CP-KGC mitigates the issue of polysemy by implementing contextual constraints. Experimentation reveals that semantic compression and expansion can enhance the model performance. The performance of text-based KGC methods can be optimized even on quantized LLMs (7B).

The progression of LLMs paves the way for continued evolution of CP-KGC. Additionally, CP-KGC can augment the data quality in real-world applications. In future, we aim to optimize open-source LLMs with specialized domain data and deepen our exploration of the confluence of KGs and LLMs in scholarly research.

\section*{Acknowledgements}
This work was funded by the Key Research and Development Program of China (Grant 2022YFC3601600), the National Natural Science Foundation of China (NSFC) (Grant 61876194), the Province Natural Science Foundation of Guangdong (Grant 2024A1515011989), the Key Research and Development Program of Guangzhou (Grant 202206010028), and the Fundamental Research Funds for the Central Universities, Sun Yat-sen University (Grant 23ptpy119).

\bibliographystyle{unsrtnat}
\bibliography{references}  

\begin{thebibliography}{36}
\providecommand{\natexlab}[1]{#1}
\providecommand{\url}[1]{\texttt{#1}}
\expandafter\ifx\csname urlstyle\endcsname\relax
  \providecommand{\doi}[1]{doi: #1}\else
  \providecommand{\doi}{doi: \begingroup \urlstyle{rm}\Url}\fi

\bibitem[Sun et~al.(2019{\natexlab{a}})Sun, Bedrax-Weiss, and Cohen]{sun2019pullnet}
Haitian Sun, Tania Bedrax-Weiss, and William~W Cohen.
\newblock Pullnet: Open domain question answering with iterative retrieval on knowledge bases and text.
\newblock In \emph{Proceedings ofthe 2019 Conference on Empirical Methods in Natural Language Processing and the 9th International Joint Conference on Natural Language Processing (EMNLP-IJCNLP)}, pages 2380--2390, 2019{\natexlab{a}}.

\bibitem[Huang et~al.(2018)Huang, Zhao, Dou, Wen, and Chang]{huang2018improving}
Jin Huang, Wayne~Xin Zhao, Hongjian Dou, Ji-Rong Wen, and Edward~Y Chang.
\newblock Improving sequential recommendation with knowledge-enhanced memory networks.
\newblock In \emph{The 41st international ACM SIGIR conference on research \& development in information retrieval}, pages 505--514, 2018.

\bibitem[Bordes et~al.(2013)Bordes, Usunier, Garcia-Duran, Weston, and Yakhnenko]{bordes2013translating}
Antoine Bordes, Nicolas Usunier, Alberto Garcia-Duran, Jason Weston, and Oksana Yakhnenko.
\newblock Translating embeddings for modeling multi-relational data.
\newblock \emph{Advances in neural information processing systems}, 26, 2013.

\bibitem[Wang et~al.(2014)Wang, Zhang, Feng, and Chen]{wang2014knowledge}
Zhen Wang, Jianwen Zhang, Jianlin Feng, and Zheng Chen.
\newblock Knowledge graph embedding by translating on hyperplanes.
\newblock In \emph{Proceedings of the AAAI conference on artificial intelligence}, volume~28, 2014.

\bibitem[Sun et~al.(2019{\natexlab{b}})Sun, Deng, Nie, and Tang]{sun2019rotate}
Zhiqing Sun, Zhi-Hong Deng, Jian-Yun Nie, and Jian Tang.
\newblock Rotate: Knowledge graph embedding by relational rotation in complex space.
\newblock In \emph{7th International Conference on Learning Representations, ICLR 2019}, 2019{\natexlab{b}}.

\bibitem[Wang et~al.(2022)Wang, Zhao, Wei, and Liu]{wang2022simkgc}
Liang Wang, Wei Zhao, Zhuoyu Wei, and Jingming Liu.
\newblock Simkgc: Simple contrastive knowledge graph completion with pre-trained language models.
\newblock In \emph{Proceedings of the 60th Annual Meeting of the Association for Computational Linguistics, ACL 2022}, volume 1: Long Papers, pages 4281--4294, 2022.

\bibitem[Yao et~al.(2019)Yao, Mao, and Luo]{yao2019kg}
Liang Yao, Chengsheng Mao, and Yuan Luo.
\newblock Kg-bert: Bert for knowledge graph completion.
\newblock \emph{arXiv preprint arXiv:1909.03193}, 2019.

\bibitem[Singhal et~al.(2023)Singhal, Azizi, Tu, Mahdavi, Wei, Chung, Scales, Tanwani, Cole-Lewis, Pfohl, et~al.]{singhal2023large}
Karan Singhal, Shekoofeh Azizi, Tao Tu, S.~Sara Mahdavi, Jason Wei, Hyung~Won Chung, Nathan Scales, Ajay Tanwani, Heather Cole-Lewis, Stephen Pfohl, et~al.
\newblock Large language models encode clinical knowledge.
\newblock \emph{Nature}, 620\penalty0 (7972):\penalty0 172--180, 2023.
\newblock \doi{10.1038/s41586-023-06291-2}.

\bibitem[Petroni et~al.(2019)Petroni, Rockt{\"a}schel, Lewis, Bakhtin, Wu, Miller, and Riedel]{petroni2019language}
Fabio Petroni, Tim Rockt{\"a}schel, Patrick Lewis, Anton Bakhtin, Yuxiang Wu, Alexander~H Miller, and Sebastian Riedel.
\newblock Language models as knowledge bases?
\newblock In \emph{Proceedings of the 2019 Conference on Empirical Methods in Natural Language Processing and the 9th International Joint Conference on Natural Language Processing (EMNLP-IJCNLP)}, pages 2463--2473, 2019.

\bibitem[Amatriain(2024)]{amatriain2024prompt}
Xavier Amatriain.
\newblock Prompt design and engineering: Introduction and advanced methods.
\newblock \emph{arXiv preprint arXiv:2401.14423}, 2024.

\bibitem[Bai et~al.(2023)Bai, Bai, Chu, Cui, Dang, Deng, Fan, Ge, Han, Huang, Hui, Ji, Li, Lin, Lin, Liu, Liu, Lu, Lu, Ma, Men, Ren, Ren, Tan, Tan, Tu, Wang, Wang, Wang, Wu, Xu, Xu, Yang, Yang, Yang, Yang, Yao, Yu, Yuan, Yuan, Zhang, Zhang, Zhang, Zhang, Zhou, Zhou, Zhou, and Zhu]{Jinze2023Qwen}
Jinze Bai, Shuai Bai, Yunfei Chu, Zeyu Cui, Kai Dang, Xiaodong Deng, Yang Fan, Wenbin Ge, Yu~Han, Fei Huang, Binyuan Hui, Luo Ji, Mei Li, Junyang Lin, Runji Lin, Dayiheng Liu, Gao Liu, Chengqiang Lu, Keming Lu, Jianxin Ma, Rui Men, Xingzhang Ren, Xuancheng Ren, Chuanqi Tan, Sinan Tan, Jianhong Tu, Peng Wang, Shijie Wang, Wei Wang, Shengguang Wu, Benfeng Xu, Jin Xu, An~Yang, Hao Yang, Jian Yang, Shusheng Yang, Yang Yao, Bowen Yu, Hongyi Yuan, Zheng Yuan, Jianwei Zhang, Xingxuan Zhang, Yichang Zhang, Zhenru Zhang, Chang Zhou, Jingren Zhou, Xiaohuan Zhou, and Tianhang Zhu.
\newblock Qwen technical report.
\newblock \emph{arXiv preprint arXiv:2309.16609}, 2023.

\bibitem[Touvron et~al.(2023)Touvron, Martin, Stone, Albert, Almahairi, Babaei, Bashlykov, Batra, Bhargava, Bhosale, et~al.]{touvron2023llama}
Hugo Touvron, Louis Martin, Kevin Stone, Peter Albert, Amjad Almahairi, Yasmine Babaei, Nikolay Bashlykov, Soumya Batra, Prajjwal Bhargava, Shruti Bhosale, et~al.
\newblock Llama 2: Open foundation and fine-tuned chat models.
\newblock \emph{arXiv preprint arXiv:2307.09288}, 2023.

\bibitem[Achiam et~al.(2023)Achiam, Adler, Agarwal, Ahmad, Akkaya, Aleman, Almeida, Altenschmidt, Altman, Anadkat, et~al.]{achiam2023gpt}
Josh Achiam, Steven Adler, Sandhini Agarwal, Lama Ahmad, Ilge Akkaya, Florencia~Leoni Aleman, Diogo Almeida, Janko Altenschmidt, Sam Altman, Shyamal Anadkat, et~al.
\newblock Gpt-4 technical report.
\newblock \emph{arXiv preprint arXiv:2303.08774}, 2023.

\bibitem[Lin et~al.(2015)Lin, Liu, Sun, Liu, and Zhu]{lin2015learning}
Yankai Lin, Zhiyuan Liu, Maosong Sun, Yang Liu, and Xuan Zhu.
\newblock Learning entity and relation embeddings for knowledge graph completion.
\newblock In \emph{Proceedings of the AAAI conference on artificial intelligence}, volume~29, 2015.

\bibitem[Trouillon et~al.(2016)Trouillon, Welbl, Riedel, Gaussier, and Bouchard]{trouillon2016complex}
Th{\'e}o Trouillon, Johannes Welbl, Sebastian Riedel, {\'E}ric Gaussier, and Guillaume Bouchard.
\newblock Complex embeddings for simple link prediction.
\newblock In \emph{International conference on machine learning}, pages 2071--2080. PMLR, 2016.

\bibitem[Wang et~al.(2021{\natexlab{a}})Wang, Gao, Zhu, Zhang, Liu, Li, and Tang]{wang2021kepler}
Xiaozhi Wang, Tianyu Gao, Zhaocheng Zhu, Zhengyan Zhang, Zhiyuan Liu, Juanzi Li, and Jian Tang.
\newblock Kepler: A unified model for knowledge embedding and pre-trained language representation.
\newblock \emph{Transactions of the Association for Computational Linguistics}, 9:\penalty0 176--194, 2021{\natexlab{a}}.

\bibitem[Wang et~al.(2021{\natexlab{b}})Wang, Shen, Long, Zhou, Wang, and Chang]{wang2021structure}
Bo~Wang, Tao Shen, Guodong Long, Tianyi Zhou, Ying Wang, and Yi~Chang.
\newblock Structure-augmented text representation learning for efficient knowledge graph completion.
\newblock In \emph{Proceedings of the Web Conference 2021}, pages 1737--1748, 2021{\natexlab{b}}.

\bibitem[Chen et~al.(2022)Chen, Wang, Li, and Lam]{chen2022knowledge}
Chen Chen, Yufei Wang, Bing Li, and Kwok-Yan Lam.
\newblock Knowledge is flat: A seq2seq generative framework for various knowledge graph completion.
\newblock In \emph{Proceedings of the 29th International Conference on Computational Linguistics}, pages 4005--4017, 2022.

\bibitem[Chen et~al.(2023)Chen, Wang, Sun, Li, and Lam]{chen2023dipping}
Chen Chen, Yufei Wang, Aixin Sun, Bing Li, and Kwok-Yan Lam.
\newblock Dipping plms sauce: Bridging structure and text for effective knowledge graph completion via conditional soft prompting.
\newblock In \emph{Findings of the Association for Computational Linguistics: ACL 2023}, pages 11489--11503, 2023.

\bibitem[Zhao et~al.(2023)Zhao, Zhou, Li, Tang, Wang, Hou, Min, Zhang, Zhang, Dong, et~al.]{zhao2023survey}
Wayne~Xin Zhao, Kun Zhou, Junyi Li, Tianyi Tang, Xiaolei Wang, Yupeng Hou, Yingqian Min, Beichen Zhang, Junjie Zhang, Zican Dong, et~al.
\newblock A survey of large language models.
\newblock \emph{arXiv preprint arXiv:2303.18223}, 2023.

\bibitem[Wei et~al.(2022{\natexlab{a}})Wei, Tay, Bommasani, Raffel, Zoph, Borgeaud, Yogatama, Bosma, Zhou, Metzler, et~al.]{wei2022emergent}
Jason Wei, Yi~Tay, Rishi Bommasani, Colin Raffel, Barret Zoph, Sebastian Borgeaud, Dani Yogatama, Maarten Bosma, Denny Zhou, Donald Metzler, et~al.
\newblock Emergent abilities of large language models.
\newblock \emph{arXiv preprint arXiv:2206.07682}, 2022{\natexlab{a}}.

\bibitem[Pan et~al.(2024)Pan, Luo, Wang, Chen, Wang, and Wu]{pan2024unifying}
Shirui Pan, Linhao Luo, Yufei Wang, Chen Chen, Jiapu Wang, and Xindong Wu.
\newblock Unifying large language models and knowledge graphs: A roadmap.
\newblock \emph{IEEE Transactions on Knowledge and Data Engineering}, 2024.

\bibitem[Zhou et~al.(2022)Zhou, Muresanu, Han, Paster, Pitis, Chan, and Ba]{zhou2022large}
Yongchao Zhou, Andrei~Ioan Muresanu, Ziwen Han, Keiran Paster, Silviu Pitis, Harris Chan, and Jimmy Ba.
\newblock Large language models are human-level prompt engineers.
\newblock In \emph{NeurIPS 2022 Foundation Models for Decision Making Workshop}, 2022.

\bibitem[Shin et~al.(2020)Shin, Razeghi, Logan~IV, Wallace, and Singh]{shin2020autoprompt}
Taylor Shin, Yasaman Razeghi, Robert~L Logan~IV, Eric Wallace, and Sameer Singh.
\newblock Autoprompt: Eliciting knowledge from language models with automatically generated prompts.
\newblock In \emph{Proceedings of the 2020 Conference on Empirical Methods in Natural Language Processing (EMNLP)}, pages 4222--4235, 2020.

\bibitem[Brown et~al.(2020)Brown, Mann, Ryder, Subbiah, Kaplan, Dhariwal, Neelakantan, Shyam, Sastry, Askell, et~al.]{brown2020language}
Tom Brown, Benjamin Mann, Nick Ryder, Melanie Subbiah, Jared~D Kaplan, Prafulla Dhariwal, Arvind Neelakantan, Pranav Shyam, Girish Sastry, Amanda Askell, et~al.
\newblock Language models are few-shot learners.
\newblock \emph{Advances in neural information processing systems}, 33:\penalty0 1877--1901, 2020.

\bibitem[Jiang et~al.(2020)Jiang, Xu, Araki, and Neubig]{jiang2020can}
Zhengbao Jiang, Frank~F Xu, Jun Araki, and Graham Neubig.
\newblock How can we know what language models know?
\newblock \emph{Transactions of the Association for Computational Linguistics}, 8:\penalty0 423--438, 2020.

\bibitem[Wei et~al.(2022{\natexlab{b}})Wei, Wang, Schuurmans, Bosma, Xia, Chi, Le, Zhou, et~al.]{wei2022chain}
Jason Wei, Xuezhi Wang, Dale Schuurmans, Maarten Bosma, Fei Xia, Ed~Chi, Quoc~V Le, Denny Zhou, et~al.
\newblock Chain-of-thought prompting elicits reasoning in large language models.
\newblock \emph{Advances in Neural Information Processing Systems}, 35:\penalty0 24824--24837, 2022{\natexlab{b}}.

\bibitem[Vaswani et~al.(2017)Vaswani, Shazeer, Parmar, Uszkoreit, Jones, Gomez, Kaiser, and Polosukhin]{vaswani2017attention}
Ashish Vaswani, Noam Shazeer, Niki Parmar, Jakob Uszkoreit, Llion Jones, Aidan~N Gomez, {\L}ukasz Kaiser, and Illia Polosukhin.
\newblock Attention is all you need.
\newblock \emph{Advances in neural information processing systems}, 30, 2017.

\bibitem[Devlin et~al.(2018)Devlin, Chang, Lee, and Toutanova]{devlin2018bert}
Jacob Devlin, Ming-Wei Chang, Kenton Lee, and Kristina Toutanova.
\newblock Bert: Pre-training of deep bidirectional transformers for language understanding.
\newblock \emph{arXiv preprint arXiv:1810.04805}, 2018.

\bibitem[Liu et~al.(2019)Liu, Ott, Goyal, Du, Joshi, Chen, Levy, Lewis, Zettlemoyer, and Stoyanov]{liu2019roberta}
Yinhan Liu, Myle Ott, Naman Goyal, Jingfei Du, Mandar Joshi, Danqi Chen, Omer Levy, Mike Lewis, Luke Zettlemoyer, and Veselin Stoyanov.
\newblock Roberta: A robustly optimized bert pretraining approach.
\newblock \emph{arXiv preprint arXiv:1907.11692}, 2019.

\bibitem[Raffel et~al.(2020)Raffel, Shazeer, Roberts, Lee, Narang, Matena, Zhou, Li, and Liu]{raffel2020exploring}
Colin Raffel, Noam Shazeer, Adam Roberts, Katherine Lee, Sharan Narang, Michael Matena, Yanqi Zhou, Wei Li, and Peter~J Liu.
\newblock Exploring the limits of transfer learning with a unified text-to-text transformer.
\newblock \emph{Journal of machine learning research}, 21\penalty0 (140):\penalty0 1--67, 2020.

\bibitem[Bano et~al.(2023)Bano, Khalid, Tairan, Shah, and Khattak]{bano2023summarization}
Sheher Bano, Shah Khalid, Nasser~Mansoor Tairan, Habib Shah, and Hasan~Ali Khattak.
\newblock Summarization of scholarly articles using bert and bigru: Deep learning-based extractive approach.
\newblock \emph{Journal of King Saud University-Computer and Information Sciences}, 35\penalty0 (9):\penalty0 101739, 2023.

\bibitem[Bano and Khalid(2022)]{bano2022bert}
Sheher Bano and Shah Khalid.
\newblock Bert-based extractive text summarization of scholarly articles: A novel architecture.
\newblock In \emph{2022 International Conference on Artificial Intelligence of Things (ICAIoT)}, pages 1--5. IEEE, 2022.

\bibitem[Yang et~al.(2014)Yang, Yih, He, Gao, and Deng]{yang2014embedding}
Bishan Yang, Wen-tau Yih, Xiaodong He, Jianfeng Gao, and Li~Deng.
\newblock Embedding entities and relations for learning and inference in knowledge bases.
\newblock In \emph{3rd International Conference on Learning Representations, ICLR 2015}, 2014.

\bibitem[Dettmers et~al.(2018)Dettmers, Minervini, Stenetorp, and Riedel]{dettmers2018convolutional}
Tim Dettmers, Pasquale Minervini, Pontus Stenetorp, and Sebastian Riedel.
\newblock Convolutional 2d knowledge graph embeddings.
\newblock In \emph{Proceedings of the AAAI conference on artificial intelligence}, volume~32, 2018.

\bibitem[Kaplan et~al.(2020)Kaplan, McCandlish, Henighan, Brown, Chess, Child, Gray, Radford, Wu, and Amodei]{kaplan2020scaling}
Jared Kaplan, Sam McCandlish, Tom Henighan, Tom~B Brown, Benjamin Chess, Rewon Child, Scott Gray, Alec Radford, Jeffrey Wu, and Dario Amodei.
\newblock Scaling laws for neural language models.
\newblock \emph{arXiv preprint arXiv:2001.08361}, 2020.

\end{thebibliography}






\end{document}